\documentclass{article}
\usepackage[final]{neurips_2024_ml4ps}
\usepackage[utf8]{inputenc} % allow utf-8 input
\usepackage[T1]{fontenc}    % use 8-bit T1 fonts
\usepackage{hyperref}       % hyperlinks
\usepackage{url}            % simple URL typesetting
\usepackage{booktabs}       % professional-quality tables
\usepackage{amsfonts}       % blackboard math symbols
\usepackage{nicefrac}       % compact symbols for 1/2, etc.
\usepackage{microtype}      % microtypography
\usepackage{xcolor}         % colors

\usepackage{graphicx}
\usepackage{subfig}
\usepackage{amsmath}

\global\long\def\UPCBRACE#1#2{\overset{{\scriptstyle #2}}{\overbrace{#1}}}
\global\long\def\DOWNCBRACE#1#2{\underset{{\scriptstyle #2}}{\underbrace{#1}}}

\begin{document}

\title{Exact and approximate error bounds for physics-informed neural networks}

\author{%
Augusto T. Chantada$^{1,2,}$\thanks{Correspondence to \href{mailto:augustochantada01@gmail.com}{\texttt{augustochantada01@gmail.com}}} \quad Pavlos Protopapas$^1$ \quad Luca Gomez Bachar$^{3}$\\
\textbf{Susana J. Landau}$^4$ \quad \textbf{Claudia G. Scóccola}$^5$\\
\small
$^1$John A. Paulson School of Engineering and Applied Sciences, Harvard University, \\ \small Cambridge, Massachusetts 02138, USA \\ 
\small$^2$Departamento de Física, Facultad de Ciencias Exactas y Naturales, Universidad de Buenos Aires, \\  \small Av. Intendente Cantilo S/N 1428 Ciudad Autónoma de Buenos Aires, Argentina \\ 
\small$^3$Instituto de Tecnologías en Detección y Astropartículas (CNEA, CONICET, UNSAM),\\ \small Centro Atómico  Constituyentes, San Martín, Buenos Aires CP B1650KNA, Argentina\\
\small$^4$CONICET - Universidad de Buenos Aires, Instituto de Física de Buenos Aires (IFIBA), \\ \small Av. Intendente Cantilo S/N 1428 Ciudad Autónoma de Buenos Aires, Argentina \\
\small$^5$Cosmology and Theoretical Astrophysics Group, Departamento de Física, FCFM, Universidad de Chile,\\ \small Blanco Encalada 2008, Santiago, Chile \\
}

\maketitle
\setcounter{footnote}{0}

\begin{abstract}

The use of neural networks to solve differential equations, as an alternative to traditional numerical solvers, has increased recently.
However, error bounds for the obtained solutions have only been developed for certain equations. In this work, we report important progress in calculating error bounds of physics-informed neural networks (PINNs) solutions of nonlinear first-order ODEs. We give a general expression that describes the error of the solution that the PINN-based method provides for a nonlinear first-order ODE. In addition, we propose a technique to calculate an approximate bound for the general case and an exact bound for a particular case. The error bounds are computed using only the residual information and the equation structure. We apply the proposed methods to particular cases and show that they can successfully provide error bounds without relying on the numerical solution.

\end{abstract}

\section{Introduction}
Since its conception in \cite{Lagaris} and popularization by \cite{PINNs}, physics-informed neural networks (PINNs) are gaining popularity as an alternative to numerical methods to solve differential equations (\cite{PINNs_flows,PINNs_fluid,QE_pinns}).
The solutions provided by PINN-based methods have many advantages, such as being continuous and fully differentiable (\cite{HFM,faster_bayesian}).

However, PINNs should be able to provide error bounds on the total error of their solutions to offer a competitive alternative to numerical methods. Although the literature on error bounds for PINNs is extensive, most previous work focuses on: i) proving the existence of neural networks that guarantee an arbitrarily small generalization error, ii) bounds on generalized norms of the total error and iii) bounds on the generalization error as a function of the training error (\cite{err_Kolmogorov,gen_err_bounds, err_NS, err_semilinear}). Our interest falls somewhat near ii). Indeed, the goal of this paper is to develop error bounds that provide an \textit{a posteriori} upper limit on the absolute value of the total error of a trained PINN as a function of its inputs. Besides, we seek that the bound only depends on known quantities such as the PINN-based solution and its associated residual. These types of error bounds have been addressed in previous works for the case of linear ODEs, certain nonlinear
ODEs, and first-order linear PDEs (\cite{previous_work2,previous_work}).

In this work, we study the general case of a nonlinear first-order ODE and provide a general expression for the total error of the solution that a PINN provides, as a function of its inputs. From the latter, we develop a method to compute an approximate bound for the general case and an exact bound for a particular case. Like the error bounds provided in \cite{previous_work}, the proposed algorithms only use the residual and structure of the equation. We also apply both algorithms proposed here to specific differential equations and show that they provide reasonable bounds on the obtained solutions.

\section{Methodology}
\label{sec:methods}
Our focus in this work is  a first-order ODE of the form
\begin{equation}
\label{eq:semi_gen_first_order_ODE}
    \dfrac{du}{dt}+f(u, t)=0 \quad \left.u(t)\right|_{t=t_0}=u_0,
\end{equation}
where $f\in C^{\infty}$ is an arbitrary differentiable scalar function. An approximate solution, $v(t)$, to this equation can be provided by a neural network trained to minimize the residual, $r(t)$, defined as
\begin{equation}
\label{eq:residual}
    \dfrac{dv}{dt}+f(v, t)=r(t).
\end{equation}
The loss function, $L$, that the neural network is trained on is
\begin{equation}
    L=\dfrac{1}{|I|}\int_{I}r(t)^2dt+\left[u_0-v(t_0)\right]^2,
\end{equation}
where $I$ is the temporal domain where the solution is of interest.

We define the total error that the neural network makes in approximating the solution as $\eta := u - v$. In order to find a bound of $|\eta(t)|$ we first need to find $\eta(t)$ based solely on $r(t)$ and $f$. By using $u=v+\eta$ in Eq.~\eqref{eq:semi_gen_first_order_ODE} and using the definition of the residual, the following differential equation is obtained
\begin{equation}
    \label{eq:eta_diff}
    \dfrac{d\eta}{dt}+f(v + \eta, t) - f(v,t) + r(t)=0 \quad \left.\eta(t)\right|_{t=t_0}=u_0 - v(t_0).
\end{equation}
Given that $\eta(t)$ is the solution of this last equation, solving it would give us a starting point to develop a bound. In Appendix \ref{apx:eta} we show that the solution to this equation can be expressed as
\begin{equation}
\label{eq:eta_sum}
    \eta(t)=\sum^{+\infty}_{j=0}\eta_j(t),
\end{equation}
with
\begin{subequations}
\label{eq:etas}
\begin{align}
\label{eq:eta_0}
    \eta_0(t)=&e^{-q(t)}\left[u_0 - v(t_0)-\int^t_{t_0}r(t^\prime)e^{q(t^\prime)}dt^\prime \right],\quad\\
\label{eq:eta_j}
    \eta_j(t)=&-e^{-q(t)}\int^t_{t_0}\left\{\sum^{j-2}_{k=-1}F_{j-k}(t^\prime) \left[\sum_{\substack{j_1+\cdots +j_{j-k}=k+1 \\ j_1,\dots,j_{j-k}\geq0}}\eta_{j_1}(t^\prime)\cdots\eta_{j_{j-k}}(t^\prime)\right]\right\}e^{q(t^\prime)}dt^\prime\quad \forall j>0,
\end{align}
\end{subequations}
where $F_n=\frac{1}{n!}\frac{\partial^n f(v+\eta,t)}{\partial v^n}|_{\eta=0}$ and $q(t)=\int^t_{t_0}F_1(t^\prime)dt^\prime$. With Eqs.~\eqref{eq:etas} we now proceed to develop bounds for $|\eta(t)|$. Due to the generality of the problem, we will develop an approximate bound for the general case and an exact bound for a particular choice of $f(u,t)$.
\subsection{An approximate bound for the general case}
\label{sec:approx_bound}
Given $\eta(t)=\sum^{+\infty}_{j=0}\eta_j(t)$ we have:
\begin{equation}
\label{eq:eta_bound}
    \lvert \eta(t) \rvert = \lvert \eta_0(t)+ \eta_1(t)+\eta_2(t)+\cdots \rvert
    \leq \lvert \eta_0(t)\rvert+\lvert \eta_1(t)\rvert+\lvert \eta_2(t)\rvert+ \cdots.
\end{equation}
One possible avenue for an error bound could be to truncate the above sum at an order $J$. This way
\begin{equation}
\label{eq:loose_bound}
    \lvert\eta(t)\rvert \leq \tilde{\mathcal{B}}_{\rm loose}(t; J):=\sum^{J}_{j=0}\lvert \eta_j(t)\rvert.
\end{equation}

Given that we are truncating the sum, a criterion should be developed to choose the order of the truncation. One property that we want from $\eta_{J}$ is that it does not add a significant amount to the sum. To gauge this, we can impose that
\begin{equation}
\label{J_error_criterion}
    |\eta_{J}(t)|<\varepsilon_{{\rm abs}, J}+\varepsilon_{{\rm rel}, J}|v(t)|\quad\forall t\in I,
\end{equation}
where $\varepsilon_{{\rm abs}, J}$ and $\varepsilon_{{\rm rel}, J}$ are the absolute and relative tolerances respectively. These tolerances are hyperparameters of the method that need to be set by the user.

Assuming $\eta_J$ obeys Eq.~\eqref{J_error_criterion}, we also want to check if the sum exhibits a convergent behavior. A possible criterion to check this could be that:
\begin{equation}
\label{J_convergence_criterion}
    \max_{t\in I}\left[|\eta_J(t)|\right]<\bar{\eta}_j\quad\forall j<J,
\end{equation}
where $\bar{\eta}_j=\frac{1}{|I|}\int_{I}|\eta_j(t^\prime)|dt^\prime$ is the mean of $|\eta_j(t)|$ over the domain. This avoids a problem that arises with the more intuitive criterion $|\eta_J(t)|<|\eta_j(t)|\;\forall t\in I  \land \forall j<J$. The problem with this criterion is that $\eta_j(t)$ could cross the horizontal axis and change sign along the domain, and therefore it is zero at some point in the domain, making it impossible for any order $J$ to meet the criterion.
\subsubsection{A tighter bound}
A drawback to using Eq.~\eqref{eq:loose_bound} is that it can yield bounds that are too loose. Depending on the problem and the residuals, if the approximation of the error needs a lot of orders to correctly approximate the true error, then adding the absolute value of each $\eta_j(t)$ is going to reach values far above the true error. A possible alternative that would yield a tighter bound on the error, at the possible cost of requiring more orders, is to do an intermediate step in Eq.~\eqref{eq:eta_bound} as
\begin{equation}
\begin{split}
\label{eq:eta_bound_tight}
    \lvert \eta(t) \rvert &= \lvert \eta_0(t)+ \eta_1(t)+\eta_2(t)\cdots \rvert\\
    &\leq \lvert \eta_0(t)+ \eta_1(t) + \eta_2(t)+ \cdots + \eta_P(t)\rvert + \lvert \eta_{P+1}(t)\rvert + \lvert \eta_{P+2}(t)\rvert + \cdots. \\
\end{split}
\end{equation}
Truncating this last sum at an order $J$ we obtain
\begin{equation}
\label{eq:tight_bound}
    \lvert\eta(t)\rvert \leq \tilde{\mathcal{B}}_{\rm tight}(t; P, J)=\left\lvert\sum^P_{j=0} \eta_j(t)\right\rvert + \sum^{J}_{j=P+1} \lvert\eta_j(t)\rvert.
\end{equation}
The criterion for choosing $P$ and $J$ could be the same as the loose bound, but with two different pairs of tolerances $(\varepsilon_{{\rm abs}, P}, \varepsilon_{{\rm rel}, P})$ and $(\varepsilon_{{\rm abs}, J}, \varepsilon_{{\rm rel}, J})$, such that $\varepsilon_{{\rm abs},P} > \varepsilon_{{\rm abs},J}$ and $\varepsilon_{{\rm rel},P} > \varepsilon_{{\rm rel},J}$.

The idea behind Eq.~\eqref{eq:tight_bound} is that if $\sum^P_{j=0} \eta_j(t)$ approximates $\eta(t)$ well enough, then $\sum^{J}_{j=P+1} \lvert\eta_j(t)\rvert$ is not going to add too much and blow the bound out of proportion. Basically letting the sum sort of converge and then add a few extra orders just to be safe. It can be seen from Eqs.~\eqref{eq:loose_bound} and \eqref{eq:tight_bound} that $\tilde{\mathcal{B}}_{\rm tight}(t;0,J)=\tilde{\mathcal{B}}_{\rm loose}(t;J)$. It is important to emphasize that neither bound is guaranteed not to underestimate the absolute value of the error given that these are approximate bounds.
\subsection{An exact bound for a particular case}
\label{sec:exact_bound}
Given the generality of Eq.~\eqref{eq:semi_gen_first_order_ODE}, it is difficult to develop an exact bound for $|\eta(t)|$. Nevertheless, there is a less general case for which an exact bound can be found. If we take $f(u,t)=C(t)u^2+B(t)u+A(t)$ in Eq.~\eqref{eq:semi_gen_first_order_ODE}, with $A(t)$, $B(t)$ and $C(t)$ general scalar functions of $t$, we have
\begin{equation}
\label{eq:Riccati}
    \dfrac{du}{dt}+C(t)u^2+B(t)u+A(t)=0 \quad \left.u(t)\right|_{t=t_0}=u_0.
\end{equation}
This particular case of differential equation is sometimes referred to as the Riccati equation. We can apply the methodology described at the start of this section [Eqs.~\eqref{eq:eta_sum} and \eqref{eq:etas}] to Eq.~\eqref{eq:Riccati} where, since $F_n=0 \;\forall n>2$, the only nonzero term in the sum over $k$ in Eq.~\eqref{eq:eta_j} is $k=j-2$. Also, $q(t)=\int^t_{t_0}\left[2C(t^\prime)v(t^\prime)+B(t^\prime)\right]dt^\prime$ and $r(t)=\frac{dv}{dt}+C(t)v^2+B(t)v+A(t)$. Using the expression for $\eta_j(t)$ obtained through this method we can get the following exact bound on $|\eta(t)|$ as
\begin{equation}
    \label{eq:Riccati_exact_bound}
    |\eta(t)|\leq \mathcal{B}(t; J):= \left|\sum^J_{j=0}\eta_j(t)\right|+\dfrac{R\left[RK(t-t_0)\right]^{J+1}e^{-q_\downarrow(t)}}{1-RK(t-t_0)},
\end{equation}
where $R=\max_{t^*\in I}{\left\{\left|u_0 - v(t_0)\right|e^{-q_{\uparrow}(t^*)}+\int^{t^*}_{t_0}\left|r(t^\prime)\right|e^{q_{\downarrow}(t^\prime)}dt^\prime\right\}}$, $J\in \mathbb{Z}_{\geq0}$ is a free parameter of the bound, and $K=\max_{t^*\in I}\left\{\left|C(t^*)\right|e^{-q_{\downarrow}(t^*)}\right\}$. Furthermore, two monotonic functions are defined as $q_{\uparrow}(t)=\int^t_{t_0}\max\left\{0,\mathrm{Re}\left[F_1(t^\prime)\right]\right\}dt^\prime$ and $q_{\downarrow}(t)=\int^t_{t_0}\min\left\{0,\mathrm{Re}\left[F_1(t^\prime)\right]\right\}dt^\prime$, with $F_1(t)$ for this particular case being $F_1(t)=2C(t)v(t)+B(t)$. Equation \eqref{eq:Riccati_exact_bound} is only valid if $RK(t-t_0)<1$. Not only is this result exact, but it is also arbitrarily tight as we can see that $\lim_{J\xrightarrow{}+\infty}\mathcal{B}(t;J)=|\eta(t)|$. One can alternatively obtain $J$ by imposing that the second term of Eq.~\eqref{eq:Riccati_exact_bound} be bellow some tolerance $\varepsilon>0\;\forall t\in I$. This way
\begin{equation}
\label{eq:J}
    J(\varepsilon) = \left\lceil\max_{t^*\in I}\left\{ \dfrac{\ln\left\{\varepsilon R^{-1}\left[1-RK\left(t^*-t_0\right)\right]e^{q_\downarrow(t^*)}\right\}}{\ln\left[RK\left(t^*-t_0\right)\right]}\right\}-1\right\rceil,
\end{equation}
where $\lceil \cdot\rceil$ denotes the ceiling operator and for the equation to be valid $RK(t-t_0)<1 \;\forall t\in I$. A detailed explanation of how we got to Eqs.~\eqref{eq:Riccati_exact_bound} and \eqref{eq:J} can be found in Appendix \ref{apx:exact_bound}.
\section{Results}
\label{sec:results}
Here we apply the methods described in Sec.~\ref{sec:methods} to PINNs that have been trained to solve two examples of differential equations. To train the PINNs we used the NeuroDiffeq library (\cite{neurodiff}). Also, to obtain the bounds one needs to compute integrals, as evident by Eqs.~\eqref{eq:etas}. These integrals were computed using the trapezoidal rule with $10^4$ integration points.

\begin{figure}[htp]
    \centering
    \subfloat[Approximate bounds for Eq.~\eqref{eq:pop}]{%
        \includegraphics[width=0.5\textwidth]{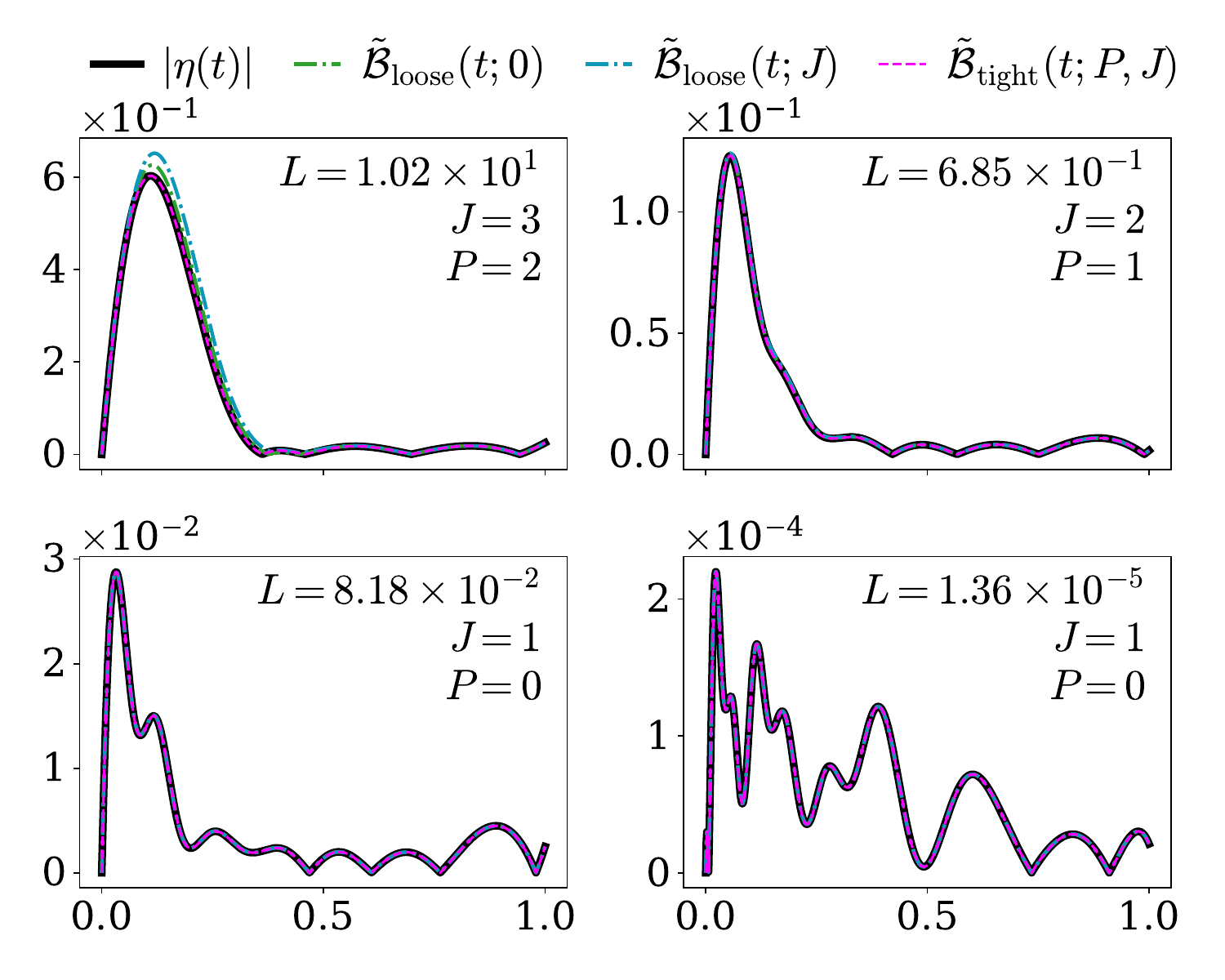}%
        \label{fig:a}%
        }
    \hfill%
    \subfloat[Exact bounds for Eq.~\eqref{eq:cosmo}]{%
        \includegraphics[width=0.5\textwidth]{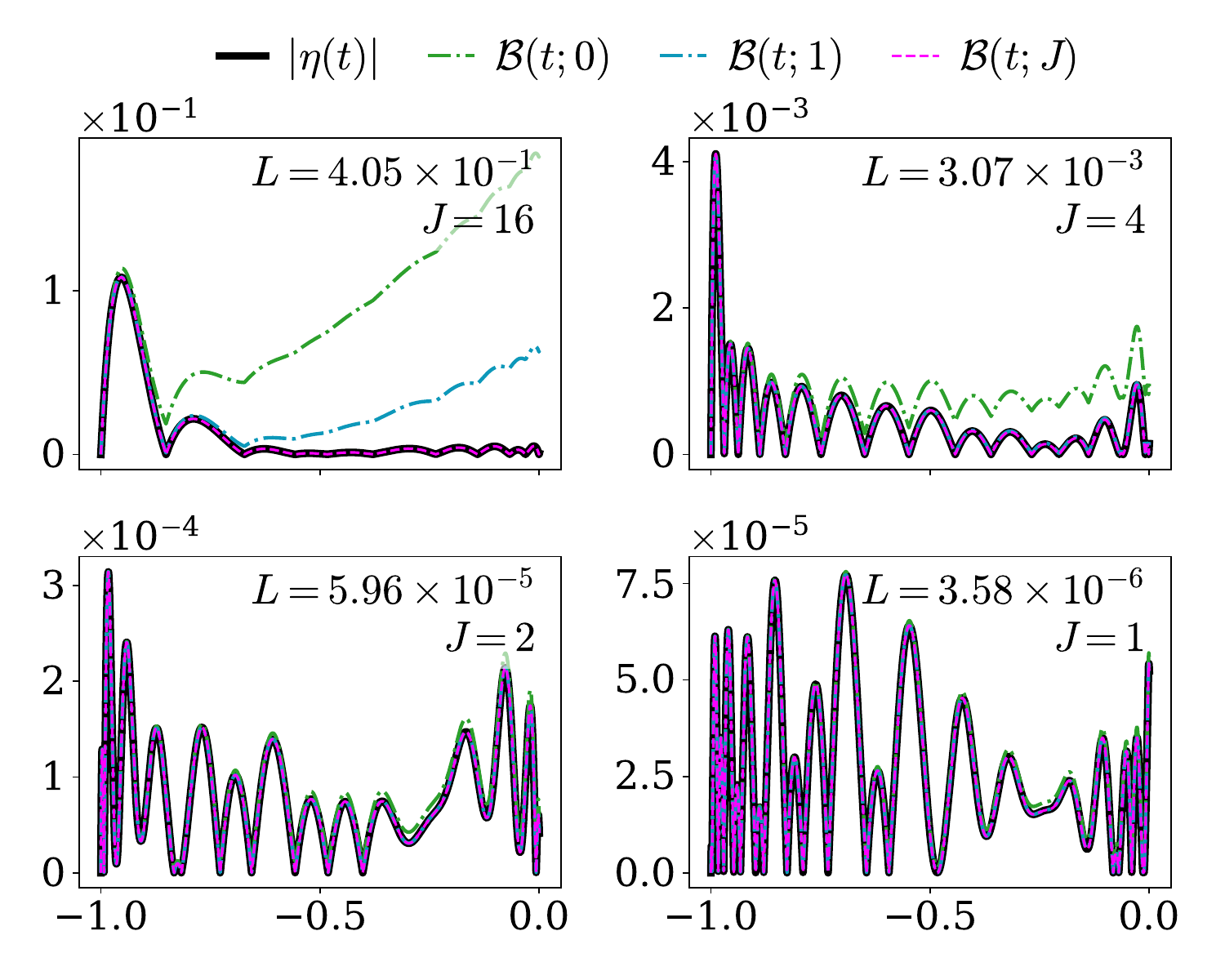}%
        \label{fig:b}%
        }%
    \caption{The absolute value of the error of PINNs with progressively lower loss, $L$, against approximate bounds for Eq.~\eqref{eq:pop} (a) and exact bounds for Eq.~\eqref{eq:cosmo} (b).}
\end{figure}

First, we test both approaches to an approximate bound with the following differential equation:
\begin{equation}
\label{eq:pop}
    \dfrac{du}{dt} -Tu\left(1 - \dfrac{u}{k} \right) + \alpha u + \dfrac{\beta u}{1+u^2} = 0,\quad \left.u(t)\right|_{t=t_0}=u_0.
\end{equation}
This equation is a nonlinear differential equation where the nonlinearity is not a polynomial, which allows us to test the limits of using a finite number of terms in the expansion of $f(v+\eta, t)$ around $\eta\sim 0$. We show in Fig.~\ref{fig:a} the absolute value of the error\footnote{The absolute value of the error is always computed using a solution given by a numerical method (RK45) with low error tolerances ($10^{-11}$ for the relative tolerance and $10^{-16}$ for the absolute tolerance).} of PINNs trained to solve Eq.~\eqref{eq:pop}, each with progressively lower loss, against different possible approximate bounds. For these we set $\varepsilon_{{\rm abs}, P}=10^{-6}, \varepsilon_{{\rm rel}, P}=10^{-3}$ and $\varepsilon_{{\rm abs}, J}=10^{-7}, \varepsilon_{{\rm rel}, J}=10^{-4}$. The parameters of Eq.~\eqref{eq:pop} were set at $T=10$, $k=5$, $\alpha = 2$, $\beta=-40$, $u_0=2$ and $t_0=0$; also the training domain was $I=[0,1]$. Figure \ref{fig:a} shows that, at least for this example, even the error of solutions with high loss and high error can be accurately bounded with low values of $P$ and $J$ for both the loose and tight bounds. Furthermore, just $P=0$, making the loose and tight bounds equivalent, and $J=1$ are enough for solutions with reasonably low losses.

To test the proposed method to calculate an exact bound, we use the following differential equation which is widely used in the cosmological context (\cite{matter_perturbations}):\footnote{This equation is used to describe the growth of structures like galaxies and clusters of galaxies in the cosmological context. The particular expression used in this work arises after making certain variable changes to the more canonical version of the equation.}
\begin{equation}
\label{eq:cosmo}
    \dfrac{du}{dt} + u^2 + \dfrac{u_0u\left(1+4\beta\gamma e^{3u_0 t}\right)}{2\left[1+\beta e^{-u_0t}\left(1 + \gamma e^{4u_0t}\right)\right]} + \dfrac{3u_0^2 e^{u_0 t}\left[1+g\left(1-e^{u_0 t}\right)^n-g\left(1-e^{u_0 t}\right)^{2n}\right]}{2\left[\beta + e^{u_0t}\left(1+\beta\gamma e^{4u_0t}\right)\right]}=0.
\end{equation}
In Fig.~\ref{fig:b} we show the absolute value of the error against the exact bounds with different values of $J$ [specifically $J=0$, $J=1$ and $J(\varepsilon=10^{-8})$ using Eq.~\eqref{eq:J}] for PINNs with progressively lower loss. The parameters of Eq.~\eqref{eq:cosmo} were set as follows: $u_0=6.91$, $\gamma=1.47\times 10^{4}$, $\beta = 2.56\times10^{-4}$, $g=-1.16$, $n=2$ and $t_0=-1$; also the training domain was $I=[-1,0]$. Similarly to the approximate bounds, Figure \ref{fig:b} shows that for the exact bounds, one does not need a high value of $J$ to get a useful bound given that the loss is low enough.

\section{Conclusions and future work}
In this work, we tackled the calculation of the total error of a PINN-based solution, for the most general case of a first-order nonlinear ODE. Based on the latter, we developed a method to compute an approximate error bound for a general case and an exact bound for a particular ODE. Finally, we applied the proposed methods to example differential equations and obtained reasonable error bounds. Our work is the first of its kind in developing bounds for general nonlinear ODEs.
We leave for future work the computation of an exact error bound of a general nonlinear first-order ODE, as well as the extension of the approximate error bound to higher-order ODEs and systems of ODEs.

\appendix
\section{Solving for \texorpdfstring{$\eta(t)$}{eta(t)}}
\label{apx:eta}
To develop the bounds of this work we first needed to solve Eq.~\eqref{eq:eta_diff} and get an expression for $\eta(t)$. Here we show how we solve it. Starting with Eq.~\eqref{eq:eta_diff} we have
\begin{equation}
\begin{split}
    \dfrac{d\eta}{dt}+f(v + \eta, t) - f(v,t) + r(t)=0,&\\
    \dfrac{d\eta}{dt}+ \sum^{+\infty}_{n=1}\dfrac{1}{n!}\left.\dfrac{\partial^n f}{\partial v^n}\right|_{\eta=0}\eta^n =-r(t),&\\
    \dfrac{d\eta}{dt}+ \left.\dfrac{\partial f}{\partial v}\right|_{\eta=0}\eta +\sum^{+\infty}_{n=2}\dfrac{1}{n!}\left.\dfrac{\partial^n f}{\partial v^n}\right|_{\eta=0}\eta^{n} =-r(t)\\
    \left[\dfrac{d}{dt}+ F_1(t)\right]\eta +\sum^{+\infty}_{n=2}F_n(t)\eta^{ n} =-r(t)&.
\end{split}
\end{equation}
Setting $\eta(t)=\sum^{+\infty}_{j=0}\eta_j(t)$, we need to first deal with the initial condition of Eq.~\eqref{eq:eta_diff}. There are infinite ways to constrain $\eta_j(t)$ such that $\eta(t_0)=u_0-v(t_0)$. We set $\eta_0(t_0)=u_0-v(t_0)$ and $\eta_j(t_0)=0\;\forall\;j>0$. Now, we deal with the differential equation:

\begin{equation}
\begin{split}
    \left[\dfrac{d}{dt}+F_1(t)\right]\sum^{+\infty}_{j=0}\eta_j(t)+\sum^{+\infty}_{n=2} F_{n}(t) \left[\sum^{+\infty}_{j=0}\eta_j(t)\right]^{n}=-r(t)&\\
    \sum^{+\infty}_{j=0}\left[\dfrac{d}{dt}+F_1(t)\right]\eta_j(t)+\sum^{+\infty}_{n=2} F_{n}(t) \left[\sum^{+\infty}_{j=0}\UPCBRACE{\sum_{\substack{j_1+\cdots +j_{n}=j \\ j_1,\dots,j_{n}\geq0}}\eta_{j_1}(t)\cdots\eta_{j_{n}}(t)}{S_{n,j}(t):=}\right]=-r(t)&\\
    \sum^{+\infty}_{j=0}\left[\dfrac{d}{dt}+F_1(t)\right]\eta_j(t)+\sum^{+\infty}_{n=2}\sum^{+\infty}_{j=0} F_{n}(t) S_{n,j}(t)=-r(t)&\\
    \left[\dfrac{d}{dt}+F_1(t)\right]\eta_0(t)+r(t)+\sum^{+\infty}_{j=1}\left[\dfrac{d}{dt}+F_1(t)\right]\eta_j(t)+\sum^{+\infty}_{n=2} \sum^{+\infty}_{j=0}F_{n}(t) S_{n,j}(t)=0&\\
    \DOWNCBRACE{\left[\dfrac{d}{dt}+F_1(t)\right]\eta_0(t)+r(t)}{=0}+\DOWNCBRACE{\sum^{+\infty}_{j=1}\left[\dfrac{d}{dt}+F_1(t)\right]\eta_j(t)+\sum^{+\infty}_{n=2} \sum^{+\infty}_{j=1}F_{n}(t) S_{n,j-1}(t)}{=0}=0
\end{split}
\end{equation}
Up until the last step, there was not any condition on the global behavior of each $\eta_j(t)$. In the last step, a particular choice of $\eta_0(t)$ was made, along with how the other $\eta_j(t)$ relate to it and each other. Therefore, for this choice, $\eta_0(t)$ is what we show in Eq.~\eqref{eq:eta_0}. Now we develop the other half of the equation further. We could take each term in the sum over $j$ to be zero, but that would require infinite terms in $n$ for each $j$. An alternative to this is to rearrange the sum so that it takes a finite number of terms for each $\eta_j$. Starting with
\begin{equation}
    \sum^{+\infty}_{j=1}\left[\dfrac{d}{dt}+F_1(t)\right]\eta_j(t)+\sum^{+\infty}_{n=2} \sum^{+\infty}_{j=1}F_{n}(t) S_{n,j-1}(t)=0,
\end{equation}
changing the mute variable $j$ in the second term to $k$
\begin{equation}
    \sum^{+\infty}_{j=1}\left[\dfrac{d}{dt}+F_1(t)\right]\eta_j(t)+\sum^{+\infty}_{n=2} \sum^{+\infty}_{k=1}F_{n}(t) S_{n,k-1}(t)=0,
\end{equation}
and now defining $j=n+k-2\implies n=j-k+2$:
\begin{equation}
\begin{split}
    \sum^{+\infty}_{j=1}\left[\dfrac{d}{dt}+F_1(t)\right]\eta_j(t)+\sum^{+\infty}_{j=1} \sum^{j}_{k=1}F_{j-k+2}(t) S_{j-k+2,k-1}(t)=0&\\
\sum^{+\infty}_{j=1}\left\{\left[\dfrac{d}{dt}+F_1(t)\right]\eta_j(t)+ \sum^{j}_{k=1}F_{j-k+2}(t) S_{j-k+2,k-1}(t)\right\}=0&\\
\sum^{+\infty}_{j=1}\left\{\DOWNCBRACE{\left[\dfrac{d}{dt}+F_1(t)\right]\eta_j(t)+ \sum^{j-2}_{k=-1}F_{j-k}(t) S_{j-k,k+1}(t)}{=0}\right\}=0
\end{split}
\end{equation}
With that last step the $\eta_j(t)$'s for $j>0$ are defined, and their expression is as shown in Eq.~\eqref{eq:eta_j}.
\section{Obtaining an exact bound}
\label{apx:exact_bound}
In this appendix, we show the calculations that led to Eqs.~\eqref{eq:Riccati_exact_bound} and \eqref{eq:J}. Before developing the bound for $|\eta(t)|$, it is useful to focus on $q(t)=\int^t_{t_0}F_1(t^\prime)dt^\prime$, where we recall that $F_n=\frac{1}{n!}\frac{\partial^n f(v+\eta,t)}{\partial v^n}|_{\eta=0}$:
\begin{equation}
\label{eq:q_down_up}
\begin{split}
    q(t)&=\int^t_{t_0}F_1(t^\prime)dt^\prime\\
        &=\mathrm{Re}\left[\int^t_{t_0}F_1(t^\prime)dt^\prime\right] + i\mathrm{Im}\left[\int^t_{t_0}F_1(t^\prime)dt^\prime\right]\\
        &=\int^t_{t_0}\mathrm{Re}\left[F_1(t^\prime)\right]dt^\prime + i\mathrm{Im}\left[q(t)\right]\\
        &=\int^t_{t_0}\left\{\max\left\{0,\mathrm{Re}\left[F_1(t^\prime)\right]\right\}+\min\left\{0,\mathrm{Re}\left[F_1(t^\prime)\right]\right\}\right\}dt^\prime + i\mathrm{Im}\left[q(t)\right]\\
        &=\DOWNCBRACE{\int^t_{t_0}\max\left\{0,\mathrm{Re}\left[F_1(t^\prime)\right]\right\}dt^\prime}{q_{\uparrow}(t):=} + \DOWNCBRACE{\int^t_{t_0}\min\left\{0,\mathrm{Re}\left[F_1(t^\prime)\right]\right\}dt^\prime}{q_{\downarrow}(t):=} + i\mathrm{Im}\left[q(t)\right].
\end{split}
\end{equation}
The functions $q_{\uparrow}(t)$ and $q_{\downarrow}(t)$ are monotonic functions. Specifically, $q_{\uparrow}(t)$ is a monotonically non-decreasing function and $q_{\uparrow}(t)\geq 0\;\forall t$. On the contrary, $q_{\downarrow}(t)$ is a monotonically non-increasing function and $q_{\downarrow}(t)\leq 0\;\forall t$. These results stem from the fact that these are integrals of quantities that are larger or equal to zero for $q_{\uparrow}(t)$, and smaller or equal to zero for $q_{\downarrow}(t)$.

With $q_{\uparrow}(t)$ and $q_{\downarrow}(t)$ now clearly defined, the next step to develop the bound for $|\eta(t)|$ is to prove the validity of the following bound for each $|\eta_j(t)|$:

\begin{equation}
\label{eq:eta_j_bound}
    |\eta_j(t)|\leq R\left[RK(t-t_0)\right]^je^{-q_\downarrow(t)},
\end{equation}

where $R=\max_{t^*\in I}{\left\{\left|u_0 - v(t_0)\right|e^{-q_{\uparrow}(t^*)}+\int^{t^*}_{t_0}\left|r(t^\prime)\right|e^{q_{\downarrow}(t^\prime)}dt^\prime\right\}}$, with $q_\downarrow(t)$ and $q_\uparrow(t)$ as defined in Eq.~\eqref{eq:q_down_up} with $F_1(t)=2C(t)v(t)+B(t)$, and $K=\max_{t^*\in I}\left\{\left|C(t^*)\right|e^{-q_{\downarrow}(t^*)}\right\}$.

To prove Eq.~\eqref{eq:eta_j_bound} we will make use of strong induction. Thus, we must prove that Eq.~\eqref{eq:eta_j_bound} holds for $j+1$ under the assumption that it holds $\forall n\in \mathbb{Z}_{\geq0}$ less or equal to $j$. In addition, we need to prove the base case, which is that Eq.~\eqref{eq:eta_j_bound} holds for $j=0$. Starting with the base case we take the absolute value of $\eta_0(t)$ in Eq.~\eqref{eq:eta_0} to obtain:
\begin{equation}
\begin{split}
    |\eta_0(t)|&=\left|e^{-q(t)}\left[u_0 - v(t_0)-\int^t_{t_0}r(t^\prime)e^{q(t^\prime)}dt^\prime\right]\right|\\
    &=\left|e^{-i\mathrm{Im}\left[q(t)\right]}\right|\left|e^{-\left[q_{\uparrow}(t)+q_{\downarrow}(t)\right]}\left[u_0 - v(t_0)-\int^t_{t_0}r(t^\prime)e^{q(t^\prime)}dt^\prime\right]\right|\\
    &=\left|e^{-q_{\downarrow}(t)}\left\{\left[u_0 - v(t_0)\right]e^{-q_{\uparrow}(t)}-\int^t_{t_0}r(t^\prime)e^{q_{\downarrow}(t^\prime)+i\mathrm{Im}\left[q(t^\prime)\right]}e^{q_{\uparrow}(t^\prime)-q_{\uparrow}(t)}dt^\prime\right\}\right|\\
    &\leq e^{-q_{\downarrow}(t)}\left\{\left|u_0 - v(t_0)\right|e^{-q_{\uparrow}(t)}+\int^t_{t_0}\left|r(t^\prime)\right|e^{q_{\downarrow}(t^\prime)}\UPCBRACE{\left|e^{i\mathrm{Im}\left[q(t^\prime)\right]}\right|}{=1}\UPCBRACE{e^{q_{\uparrow}(t^\prime)-q_{\uparrow}(t)}}{\leq 1}dt^\prime\right\}\\
\end{split}
\end{equation}
where the fact that $t\geq t^\prime\geq t_0\;\forall t\in I\implies q_{\uparrow}(t)\geq q_{\uparrow}(t^\prime)$ was used in the last step. Then
\begin{equation}
\label{eq:eta_0_bound}
    |\eta_0(t)|\leq Re^{-q_{\downarrow}(t)}= R\left[RK(t-t_0)\right]^0e^{-q_\downarrow(t)},
\end{equation}
thus proving that Eq.~\eqref{eq:eta_j_bound} holds for the base case $j=0$.

Proceeding with the other part of the proof, we start by taking the absolute value of $|\eta_{j+1}(t)|$ using Eq.~\eqref{eq:eta_j}:

\begin{equation}
\begin{split}
    |\eta_{j+1}(t)|&=\left|e^{-q(t)}\int^t_{t_0}\left\{\sum^{j-1}_{k=-1}F_{j+1-k}(t^\prime) \left[\sum_{\substack{j_1+\cdots +j_{j+1-k}=k+1 \\ j_1,\dots,j_{j+1-k}\geq0}}\eta_{j_1}(t^\prime)\cdots\eta_{j_{j+1-k}}(t^\prime)\right]\right\}e^{q(t^\prime)}dt^\prime\right|\\
    &=\left|e^{-q(t)}\int^t_{t_0}C(t^\prime) \left[\sum_{\substack{j_1+j_{2}=j \\ j_1,j_{2}\geq0}}\eta_{j_1}(t^\prime)\eta_{j_{2}}(t^\prime)\right]e^{q(t^\prime)}dt^\prime\right|\\
    &=\left|e^{-i\mathrm{Im}\left[q(t)\right]}\right|\left|e^{-\left[q_{\uparrow}(t)+q_{\downarrow}(t)\right]}\int^t_{t_0}C(t^\prime) \left[\sum_{\substack{j_1+j_{2}=j \\ j_1,j_{2}\geq0}}\eta_{j_1}(t^\prime)\eta_{j_{2}}(t^\prime)\right]e^{q(t^\prime)}dt^\prime\right|\\
    &=\left|e^{-q_{\downarrow}(t)}\int^t_{t_0}C(t^\prime) \left[\sum_{\substack{j_1+j_{2}=j \\ j_1,j_{2}\geq0}}\eta_{j_1}(t^\prime)\eta_{j_{2}}(t^\prime)\right]e^{q_{\downarrow}(t^\prime)+i\mathrm{Im}\left[q(t^\prime)\right]}e^{q_{\uparrow}(t^\prime)-q_{\uparrow}(t)}dt^\prime\right|\\
    &\leq e^{-q_{\downarrow}(t)}\int^t_{t_0}\left|C(t^\prime)\right|\left[\sum_{\substack{j_1+j_{2}=j \\ j_1,j_{2}\geq0}}|\eta_{j_1}(t^\prime)||\eta_{j_{2}}(t^\prime)|\right]e^{q_{\downarrow}(t^\prime)}\UPCBRACE{\left|e^{i\mathrm{Im}\left[q(t^\prime)\right]}\right|}{=1}\UPCBRACE{e^{q_{\uparrow}(t^\prime)-q_{\uparrow}(t)}}{\leq 1}dt^\prime\\
    &\leq e^{-q_{\downarrow}(t)}\int^t_{t_0}\left|C(t^\prime)\right|\left[\sum_{\substack{j_1+j_{2}=j \\ j_1,j_{2}\geq0}}|\eta_{j_1}(t^\prime)||\eta_{j_{2}}(t^\prime)|\right]e^{q_{\downarrow}(t^\prime)}dt^\prime,
\end{split}
\end{equation}
where in the first step we used the fact that for the Riccati equation $F_n(t)=0 \;\forall n>2$ and $F_2(t)=C(t)$. Given that $j_1+j_2=j$ with $j_1,j_2\geq 0$, then $j_1,j_2\leq j$, therefore we can use our assumption that the bound in Eq.~\eqref{eq:eta_j_bound} holds for any non-negative whole number lower than $j+1$, in particular for $j_1$ and $j_2$:
\begin{equation}
\begin{split}
    |\eta_{j+1}(t)|&\leq e^{-q_{\downarrow}(t)}\int^t_{t_0}\left|C(t^\prime)\right|\left\{\sum_{\substack{j_1+j_{2}=j \\ j_1,j_{2}\geq0}}R^2e^{-2q_\downarrow(t^\prime)}\left[RK(t^\prime-t_0)\right]^{j_1+j_2}\right\}e^{q_{\downarrow}(t^\prime)}dt^\prime.
\end{split}
\end{equation}
Then, we can use $j_1+j_2=j$:
\begin{equation}
\begin{split}
    |\eta_{j+1}(t)|&\leq R^2(RK)^j \left(\sum_{\substack{j_1+j_{2}=j \\ j_1,j_{2}\geq0}}1\right)e^{-q_{\downarrow}(t)}\int^t_{t_0}\left|C(t^\prime)\right|e^{-q_{\downarrow}(t^\prime)}(t^\prime-t_0)^{j}dt^\prime.
\end{split}
\end{equation}
Finally, using the fact that $\sum_{\substack{j_1+j_{2}=j \\ j_1,j_{2}\geq0}} 1=\binom{n+k-1}{k-1}|_{n=j,k=2}=\binom{j+1}{1}=j+1$, and recalling the definition $K=\max_{t^*\in I}\left\{\left|C(t^*)\right|e^{-q_{\downarrow}(t^*)}\right\}$:

\begin{equation}
\begin{split}
    |\eta_{j+1}(t)|&\leq (j+1)R\left(RK\right)^{j+1} e^{-q_{\downarrow}(t)}\int^t_{t_0}(t^\prime-t_0)^{j}dt^\prime=R\left[RK(t-t_0)\right]^{j+1} e^{-q_{\downarrow}(t)},
\end{split}
\end{equation}
which is the same as taking $j+1$ as the index in Eq.~\eqref{eq:eta_j_bound}, thus completing the proof.

With this last result, we can now develop a bound for $|\eta(t)|$:
\begin{equation}
\begin{split}
    |\eta(t)|&=\left|\sum^{+\infty}_{j=0}\eta_j(t)\right|\leq \left|\sum^{J}_{j=0}\eta_j(t)\right| + \left|\sum^{+\infty}_{j=J+1}\eta_j(t)\right|\leq \left|\sum^{J}_{j=0}\eta_j(t)\right| + \sum^{+\infty}_{j=J+1}\left|\eta_j(t)\right|\\
    &\leq \left|\sum^{J}_{j=0}\eta_j(t)\right| + \sum^{+\infty}_{j=J+1}R\left[RK(t-t_0)\right]^je^{-q_\downarrow(t)}\\
    &\leq \left|\sum^{J}_{j=0}\eta_j(t)\right| + Re^{-q_\downarrow(t)} \left[RK(t-t_0)\right]^{J+1}\sum^{+\infty}_{j=J+1}\left[RK(t-t_0)\right]^{j-(J+1)}\\
    &\leq \left|\sum^{J}_{j=0}\eta_j(t)\right| + Re^{-q_\downarrow(t)} \left[RK(t-t_0)\right]^{J+1}\sum^{+\infty}_{j=0}\left[RK(t-t_0)\right]^{j}.
\end{split}
\end{equation}
From the last result, if $RK(t-t_0)<1$, then
\begin{equation}
    |\eta(t)|\leq  \left|\sum^J_{j=0}\eta_j(t)\right|+\dfrac{R\left[RK(t-t_0)\right]^{J+1}e^{-q_\downarrow(t)}}{1-RK(t-t_0)}.
\end{equation}
Instead of just choosing a value for $J$, this value could be obtained by imposing that the second term in the bound be less than some tolerance $\varepsilon >0 \;\forall t \in I$. Then:

\begin{equation}
\begin{split}
    &\dfrac{R\left[RK(t-t_0)\right]^{J+1}e^{-q_\downarrow(t)}}{1-RK(t-t_0)}<\varepsilon\implies \left[RK(t-t_0)\right]^{J+1}<\varepsilon R^{-1} \left[1-RK(t-t_0)\right]e^{q_\downarrow(t)}\\
    &\implies J+1>\log_{RK(t-t_0)}\left\{\varepsilon R^{-1} \left[1-RK(t-t_0)\right]e^{q_\downarrow(t)}\right\},
\end{split}
\end{equation}
where in the last step $<$ changed to $>$ due to $RK(t-t_0)<1\; \forall t\in I$. Continuing with the calculation we have
\begin{equation}
    J>\dfrac{\ln\left\{\varepsilon R^{-1} \left[1-RK(t-t_0)\right]e^{q_\downarrow(t)}\right\}}{\ln\left[RK(t-t_0)\right]} - 1.
\end{equation}
If we want to make sure that this last inequality holds $\forall t\in I$, me must take the maximum of the right hand side over $t$ and take the closest positive whole number greater than it. Therefore
\begin{equation}
    J = \left\lceil\max_{t^*\in I}\left\{ \dfrac{\ln\left\{\varepsilon R^{-1}\left[1-RK\left(t^*-t_0\right)\right]e^{q_\downarrow(t^*)}\right\}}{\ln\left[RK\left(t^*-t_0\right)\right]}\right\}-1\right\rceil.
\end{equation}


\begin{thebibliography}{15}
\providecommand{\natexlab}[1]{#1}
\providecommand{\url}[1]{\texttt{#1}}
\expandafter\ifx\csname urlstyle\endcsname\relax
  \providecommand{\doi}[1]{doi: #1}\else
  \providecommand{\doi}{doi: \begingroup \urlstyle{rm}\Url}\fi

\bibitem[Cai et~al.(2021)Cai, Mao, Wang, Yin, and Karniadakis]{PINNs_fluid}
S.~Cai, Z.~Mao, Z.~Wang, M.~Yin, and G.~E. Karniadakis.
\newblock Physics-informed neural networks ({PINNs}) for fluid mechanics: a review.
\newblock \emph{Acta Mechanica Sinica}, 37\penalty0 (12):\penalty0 1727--1738, Dec 2021.
\newblock ISSN 1614-3116.
\newblock \doi{10.1007/s10409-021-01148-1}.
\newblock URL \url{https://doi.org/10.1007/s10409-021-01148-1}.

\bibitem[Chantada et~al.(2024)Chantada, Landau, Protopapas, Sc\'occola, and Garraffo]{faster_bayesian}
A.~T. Chantada, S.~J. Landau, P.~Protopapas, C.~G. Sc\'occola, and C.~Garraffo.
\newblock Faster {Bayesian} inference with neural network bundles and new results for $f({R})$ models.
\newblock \emph{Phys. Rev. D}, 109:\penalty0 123514, Jun 2024.
\newblock \doi{10.1103/PhysRevD.109.123514}.
\newblock URL \url{https://link.aps.org/doi/10.1103/PhysRevD.109.123514}.

\bibitem[Chen et~al.(2020)Chen, Sondak, Protopapas, Mattheakis, Liu, Agarwal, and Giovanni]{neurodiff}
F.~Chen, D.~Sondak, P.~Protopapas, M.~Mattheakis, S.~Liu, D.~Agarwal, and M.~D. Giovanni.
\newblock {NeuroDiffEq}: A {Python} package for solving differential equations with neural networks.
\newblock \emph{Journal of Open Source Software}, 5\penalty0 (46):\penalty0 1931, 2020.
\newblock \doi{10.21105/joss.01931}.
\newblock URL \url{https://doi.org/10.21105/joss.01931}.

\bibitem[De~Felice et~al.(2010)De~Felice, Mukohyama, and Tsujikawa]{matter_perturbations}
A.~De~Felice, S.~Mukohyama, and S.~Tsujikawa.
\newblock Density perturbations in general modified gravitational theories.
\newblock \emph{Phys. Rev. D}, 82:\penalty0 023524, Jul 2010.
\newblock \doi{10.1103/PhysRevD.82.023524}.
\newblock URL \url{https://link.aps.org/doi/10.1103/PhysRevD.82.023524}.

\bibitem[De~Ryck and Mishra(2022)]{err_Kolmogorov}
T.~De~Ryck and S.~Mishra.
\newblock Error analysis for physics-informed neural networks ({PINNs}) approximating {Kolmogorov} {PDEs}.
\newblock \emph{Advances in Computational Mathematics}, 48\penalty0 (6):\penalty0 79, Nov 2022.
\newblock ISSN 1572-9044.
\newblock \doi{10.1007/s10444-022-09985-9}.
\newblock URL \url{https://doi.org/10.1007/s10444-022-09985-9}.

\bibitem[De~Ryck et~al.(2023)De~Ryck, Jagtap, and Mishra]{err_NS}
T.~De~Ryck, A.~D. Jagtap, and S.~Mishra.
\newblock {Error estimates for physics-informed neural networks approximating the {Navier–Stokes} equations}.
\newblock \emph{IMA Journal of Numerical Analysis}, 44\penalty0 (1):\penalty0 83--119, 01 2023.
\newblock ISSN 0272-4979.
\newblock \doi{10.1093/imanum/drac085}.
\newblock URL \url{https://doi.org/10.1093/imanum/drac085}.

\bibitem[Jin et~al.(2022)Jin, Mattheakis, and Protopapas]{QE_pinns}
H.~Jin, M.~Mattheakis, and P.~Protopapas.
\newblock Physics-informed neural networks for quantum eigenvalue problems.
\newblock In \emph{2022 International Joint Conference on Neural Networks (IJCNN)}, pages 1--8, 2022.
\newblock \doi{10.1109/IJCNN55064.2022.9891944}.
\newblock URL \url{https://doi.org/10.1109/IJCNN55064.2022.9891944}.

\bibitem[Lagaris et~al.(1998)Lagaris, Likas, and Fotiadis]{Lagaris}
I.~Lagaris, A.~Likas, and D.~Fotiadis.
\newblock Artificial neural networks for solving ordinary and partial differential equations.
\newblock \emph{IEEE Transactions on Neural Networks}, 9\penalty0 (5):\penalty0 987--1000, 1998.
\newblock \doi{10.1109/72.712178}.
\newblock URL \url{https://doi.org/10.1109/72.712178}.

\bibitem[Liu et~al.(2022)Liu, Huang, and Protopapas]{previous_work2}
S.~Liu, X.~Huang, and P.~Protopapas.
\newblock Evaluating error bound for physics-informed neural networks on linear dynamical systems, 2022.
\newblock URL \url{https://arxiv.org/abs/2207.01114}.

\bibitem[Liu et~al.(2023)Liu, Huang, and Protopapas]{previous_work}
S.~Liu, X.~Huang, and P.~Protopapas.
\newblock Residual-based error bound for physics-informed neural networks.
\newblock In R.~J. Evans and I.~Shpitser, editors, \emph{Proceedings of the Thirty-Ninth Conference on Uncertainty in Artificial Intelligence}, volume 216 of \emph{Proceedings of Machine Learning Research}, pages 1284--1293. PMLR, 31 Jul--04 Aug 2023.
\newblock URL \url{https://proceedings.mlr.press/v216/liu23b.html}.

\bibitem[Lorenz et~al.(2024)Lorenz, Bacho, and Kutyniok]{err_semilinear}
B.~Lorenz, A.~Bacho, and G.~Kutyniok.
\newblock Error estimation for physics-informed neural networks approximating semilinear wave equations, 2024.
\newblock URL \url{https://arxiv.org/abs/2402.07153}.

\bibitem[Mao et~al.(2020)Mao, Jagtap, and Karniadakis]{PINNs_flows}
Z.~Mao, A.~D. Jagtap, and G.~E. Karniadakis.
\newblock Physics-informed neural networks for high-speed flows.
\newblock \emph{Computer Methods in Applied Mechanics and Engineering}, 360:\penalty0 112789, 2020.
\newblock ISSN 0045-7825.
\newblock \doi{https://doi.org/10.1016/j.cma.2019.112789}.
\newblock URL \url{https://www.sciencedirect.com/science/article/pii/S0045782519306814}.

\bibitem[Mishra and Molinaro(2022)]{gen_err_bounds}
S.~Mishra and R.~Molinaro.
\newblock {Estimates on the generalization error of physics-informed neural networks for approximating PDEs}.
\newblock \emph{IMA Journal of Numerical Analysis}, 43\penalty0 (1):\penalty0 1--43, 01 2022.
\newblock ISSN 0272-4979.
\newblock \doi{10.1093/imanum/drab093}.
\newblock URL \url{https://doi.org/10.1093/imanum/drab093}.

\bibitem[Raissi et~al.(2019)Raissi, Perdikaris, and Karniadakis]{PINNs}
M.~Raissi, P.~Perdikaris, and G.~Karniadakis.
\newblock Physics-informed neural networks: A deep learning framework for solving forward and inverse problems involving nonlinear partial differential equations.
\newblock \emph{Journal of Computational Physics}, 378:\penalty0 686--707, 2019.
\newblock ISSN 0021-9991.
\newblock \doi{https://doi.org/10.1016/j.jcp.2018.10.045}.
\newblock URL \url{https://www.sciencedirect.com/science/article/pii/S0021999118307125}.

\bibitem[Raissi et~al.(2020)Raissi, Yazdani, and Karniadakis]{HFM}
M.~Raissi, A.~Yazdani, and G.~E. Karniadakis.
\newblock Hidden fluid mechanics: Learning velocity and pressure fields from flow visualizations.
\newblock \emph{Science}, 367\penalty0 (6481):\penalty0 1026--1030, 2020.
\newblock \doi{10.1126/science.aaw4741}.
\newblock URL \url{https://www.science.org/doi/abs/10.1126/science.aaw4741}.

\end{thebibliography}
\end{document}